\documentclass{article} 
\usepackage{iclr2024_conference,times}
\iclrfinalcopy

\usepackage[utf8]{inputenc} 

\usepackage[T1]{fontenc}    
\usepackage{hyperref}       
\usepackage{url}            
\usepackage{booktabs}       
\usepackage{amsfonts}       
\usepackage{nicefrac}       
\usepackage{microtype}      
\usepackage{xcolor}         
\usepackage{wrapfig}
\usepackage{graphicx}
\usepackage{amsmath,amsfonts,amsthm}       
\usepackage{mathtools}      
\usepackage{multirow}
\usepackage{diagbox}
\usepackage{array}
\PassOptionsToPackage{hyphens}{url}
\usepackage{hyperref}
\usepackage[capitalise]{cleveref}
\usepackage{comment}
\usepackage{etoolbox}
\usepackage{graphicx}
\usepackage{afterpage}
\usepackage{placeins}

\usepackage[font=small]{caption}
\usepackage{algorithm} 
\usepackage{algpseudocode}  

\usepackage{tabularx} 
\usepackage{authblk} 

\usepackage{tikz}
\usepackage{scalerel}
\usetikzlibrary{patterns}
\usepackage{bm}

\usepackage{graphicx}
\usepackage{booktabs} 
\usepackage{multirow} 
\usepackage{wrapfig}
\usepackage{subcaption}
\usepackage{svg}
\usepackage{ulem}

\title{ClimateLLM: Efficient Weather Forecasting via Frequency-Aware Large Language Models}
\author[1*]{\textbf{Shixuan Li}}
\author[1*]{\textbf{Wei Yang}}
\author[1]{\textbf{Peiyu Zhang}}
\author[1]{\textbf{Xiongye Xiao}}
\author[1]{\textbf{Defu Cao}}
\author[1]{\textbf{Yuehan Qin}}
\author[1]{\textbf{Xiaole Zhang}}
\author[1]{\textbf{Yue Zhao}}
\author[1†]{\textbf{Paul Bogdan}}

\affil[1]{\centering University of Southern California, Los Angeles, CA 90089, USA}

\begin{document}
\maketitle

\begingroup
\renewcommand\thefootnote{}
\footnotetext{\textsuperscript{*}Equal Contribution. \textsuperscript{†}Corresponding to: \texttt{pbogdan@usc.edu}.}
\endgroup

\begin{abstract}

Weather forecasting is crucial for public safety, disaster prevention and mitigation, agricultural production, and energy management, with global relevance. 
Although deep learning has significantly advanced weather prediction, current methods face critical limitations: (\textit{i}) they often struggle to capture both dynamic temporal dependencies and short-term abrupt changes, making extreme weather modeling difficult; 
(\textit{ii}) they incur high computational costs due to extensive training and resource requirements; 
(\textit{iii}) they have limited adaptability to multi-scale frequencies, leading to challenges when separating global trends from local fluctuations. 
To address these issues, we propose \textbf{ClimateLLM}, a foundation model for weather forecasting. 
It captures spatiotemporal dependencies via a cross-temporal and cross-spatial collaborative modeling framework that integrates \textit{Fourier-based frequency decomposition} with \textit{Large Language Models (LLMs)} to strengthen spatial and temporal modeling. Our framework uses a \textit{Mixture-of-Experts (MoE) mechanism} that adaptively processes different frequency components, enabling efficient handling of both global signals and localized extreme events.
In addition, we introduce a cross-temporal and cross-spatial dynamic prompting mechanism, allowing LLMs to incorporate meteorological patterns across multiple scales effectively. Extensive experiments on real-world datasets show that ClimateLLM outperforms state-of-the-art approaches in accuracy and efficiency, as a scalable solution for global weather forecasting.
 
\end{abstract}

\allowdisplaybreaks


\section{Introduction}

For almost half a century, numerical weather prediction (NWP) methods that rely on solving atmospheric partial differential equations have formed the backbone of operational forecasting~\cite{Kalnay_2002, LYNCH20083431, bauer2015quiet, nguyen2024climatelearn}. 
More recently, deep learning techniques have shown significant promise as complementary or alternative tools. 
By learning complex atmospheric patterns from large-scale data, they can sometimes outperform or supplement traditional NWP models without explicitly solving physical equations \cite{pathak2022fourcastnet, bi2023accurate, lam2023learning, price2025probabilistic, verma2024climode}. Benchmarks such as WeatherBench~\cite{rasp2024weatherbench} have standardized data formats and metrics, facilitating direct comparisons across models and promoting reproducible research. Innovative approaches include neural diffusion equations \cite{hwang2021climate}, Climax \cite{nguyen2023climax}, and FourCastNet \cite{pathak2022fourcastnet}, each demonstrating distinct ways to capture atmospheric complexity using neural networks or transformers.

Despite these advances, substantial challenges remain particularly in forecasting rare but disruptive events.
First, many deep learning models demand significant computational resources and long training periods, which limits their practical use in operational settings.
Second, extreme weather events appear infrequently in historical records, creating an imbalanced data distribution that makes accurate modeling difficult \cite{he2009learning}. This problem becomes more complex because extreme weather events often involve unique physical mechanisms that differ markedly from typical weather patterns \cite{donat2013global}.
Third, non-local atmospheric teleconnections create additional complexity, as weather conditions in distant regions can significantly affect local weather patterns \cite{gao2024prediff}. Standard error metrics that focus on average prediction accuracy often lead to models that do not effectively capture rare extreme events.

To address these challenges, we propose \textbf{ClimateLLM}, a framework that combines frequency-domain processing, dynamic prompting, and large language models (LLMs) for enhanced weather forecasting. 
At its core, our approach uses a two-dimensional Fast Fourier Transform (2D FFT) to analyze spatial patterns in the frequency domain, which helps capture both large-scale atmospheric circulation and local weather patterns. 
Meanwhile, We introduce a Frequency Mixture-of-Experts (FMoE) module that processes different frequency components using specialized experts, with particular focus on the frequency bands associated with extreme weather events. In addition, the framework employs a meta-fusion prompt design that dynamically guides the model's attention to relevant temporal and variable-specific features, facilitating better cross-variable correlations and temporal dependencies. These components are integrated with a Generative Pre-trained Transformer (GPT) backbone, which excels at modeling long-range temporal dependencies crucial for weather evolution. Our architecture significantly reduces computational requirements through efficient parameter reuse from pre-trained models and limited parameter updates during fine-tuning, making it more practical for operational deployment compared to traditional deep learning approaches that require training all parameters from scratch. 
To maintain accurate predictions across different regions, we use a latitude-weighted training approach that adjusts for the varying significance of different geographical areas.

In summary, the main contributions of this paper are as follows.
\begin{itemize}
\item \textbf{Effectiveness}. We show that GPT-based temporal modeling
well
predict multiple meteorological variables on the ERA5 dataset, extending the applicability of LLMs to 
climate forecasting.
\item \textbf{Novelty}. We propose a frequency Mixture-of-Experts structure that adaptively learns multi-scale spatial representations, improving performance on localized extremes without sacrificing broader atmospheric accuracy.
\item \textbf{Efficiency}. We significantly reduce the computational burden by leveraging a partially fine-tuned model, making high-resolution forecasting more accessible for operational use.

\end{itemize}


\section{RELATED WORK}

\subsection{Deep Learning Based Forecasting }
Deep learning-based weather forecasting models have demonstrated significant advantages over traditional numerical methods in multiple aspects~\cite{leinonen2023latent, li2024cllmatemultimodalllmweather, salman2015weather,hewage2021deep}. FourCastNet \cite{pathak2022fourcastnet} outperforms the Integrated Forecasting System in predicting small-scale variables such as precipitation and extreme weather events while operating at a fraction of the computational cost. GraphCast \cite{lam2022graphcast}, trained on historical reanalysis data, delivers highly accurate 10-day global forecasts in under a minute, outperforming traditional numerical models on 90\% of verification targets and improving severe weather prediction. GenCast \cite{price2023gencast}, a probabilistic weather model, has also proven to be more accurate and efficient than the European Center for Medium-Range Weather Forecasts (ECMWF)'s ensemble forecast \cite{molteni1996ecmwf}. Additionally, FuXi \cite{chen2023fuxi} provides 15-day global forecasts with a 6-hour temporal resolution, matching ECMWF’s ensemble mean performance while extending the skillful forecast lead time beyond ECMWF's high-resolution forecast. Moreover, some deep learning-based time series models have achieved promising results in temporal tasks ~\citep{zhou2022fedformer,zhang2023crossformer,eldele2024tslanet,yi2024fouriergnn}.

\subsection{Large Language Model for Time-series Prediction}
Many studies demonstrate that large language models (LLMs) are highly effective in time series forecasting \cite{chang2023llm4ts,sun2024testtextprototypealigned}. TIME-LLM \cite{jin2023time} is a reprogramming framework that aligns time series data with language modalities by converting time series into text prototypes before feeding them into a frozen LLM, outperforming specialized forecasting models and excelling in few-shot and zero-shot learning. The Frozen Pretrained Transformer \cite{zhou2023fitsallpowergeneraltime} shows that pre-trained language and image models can achieve state-of-the-art results across various time series tasks. Similarly, the CALF framework \cite{liu2024calfaligningllmstime} reduces distribution discrepancies between textual and temporal data, improving LLM performance in both long- and short-term forecasting with low complexity and strong few-shot capabilities. \citet{chang2024llm4tsaligningpretrainedllms} introduced a two-stage fine-tuning strategy that integrates multi-scale temporal data into pre-trained LLMs, achieving superior representation learning and performance in few-shot scenarios. 
Many researches also have shown that LLMs can potentially assist in weather forecasting \cite{wang2024exploringlargelanguagemodels, wang2024newsforecastintegratingevent, li2024cllmatemultimodalllmweather}. \citet{li2024cllmatemultimodalllmweather} introduce CLLMate (LLM for climate), a multimodal LLM using meteorological raster data and textual event data, which highlights the potential of LLMs in climate forecasting.

\subsection{Fourier Neural Operator}
Fourier Neural Operators(FNOs) \cite{li2020ddtcdr} have recently garnered considerable attention as an effective deep learning framework for learning mappings between infinite‐dimensional function spaces, which is essential for approximating the solution operators of partial differential equations.
\citet{chen2019neuralordinarydifferentialequations} provide a continuous formulation for neural networks by modeling the evolution of hidden states as solutions to differential equations, a concept that has inspired recent advances in operator learning.
Many studies demonstrate that Fourier Neural Operators (FNOs) are highly effective for data-driven forecasting of complex physical processes. They capture the continuous evolution of weather variables—such as temperature, wind speed, and atmospheric pressure—across both spatial and temporal dimensions. \citet{pathak2022fourcastnet} applies Adaptive Fourier Neural Operator(AFNO) to learn the evolution of weather variables across both spatial and temporal domains, effectively capturing the large-scale trends as well as the fine-grained structures inherent in the weather system.
\citet{sun2023rapid} employs the FNO as a surrogate model to predict flood extents and water depths at high resolution, addressing the computational challenges associated with traditional hydrodynamic simulations. Leveraging global convolution, FNOs efficiently simulate fluid dynamics, making them ideal for long-term trend modeling and data-driven forecasting.


\section{PRELIMINARIES}

This paper proposes a general climate prediction framework based on large language models. Given a climate system, let $\mathcal{V} = \{t, u, v, ...\}$ denote the set of climate variables, where $t$ represents temperature, $u$ represents wind speed, $v$ represents humidity, etc. The climate state at time step $l$ can be represented as $X(l) \in \mathbb{R}^{|\mathcal{V}|\times M\times N}$, where $M$ and $N$ denote the dimensions of the spatial grid. Specifically, $X_{true}(l)[v,m,n] \in \mathbb{R}$ represents the ground truth value of variable $v \in \mathcal{V}$ at location $(m,n)$ at time step $l$, while $X_{pred}(l)[v,m,n] \in \mathbb{R}$ represents the predicted value. Let $\mathcal{H}(t) = \{X_{true}(t-L)[v,m,n], ..., \\X_{true}(t-1)[v,m,n]\} \in \mathbb{R}^{L\times|\mathcal{V}|\times M\times N}$ denote the historical sequence of length $L$ leading up to time $t$. A sample in our dataset can be represented as $(x_s, y_s)$, where $x_s = \mathcal{H}(t)$ represents the input features constructed from the historical sequence, and $y_s = X_{true}(t)[v,m,n]$ represents the ground truth value at the target time step. The prediction function $f$ can be formulated as: $f: \mathbb{R}^{L\times|\mathcal{V}|\times M\times N} \rightarrow \mathbb{R}^{|\mathcal{V}|\times M\times N}$ where $f(\mathcal{H}(t)) = X_{pred}(t)$ represents the predicted climate state at time $t$. This paper mainly studies the climate prediction problem, which is to learn the optimal prediction function $f^*$ that minimizes the prediction error: $f^* = \arg\min_f \mathcal{L}_{RMSE}(X_{pred}(t), X_{true}(t))$.


\section{THE PROPOSED MODEL}

\begin{figure*}[h]
\small
  \centering
  \includegraphics[width=\linewidth]{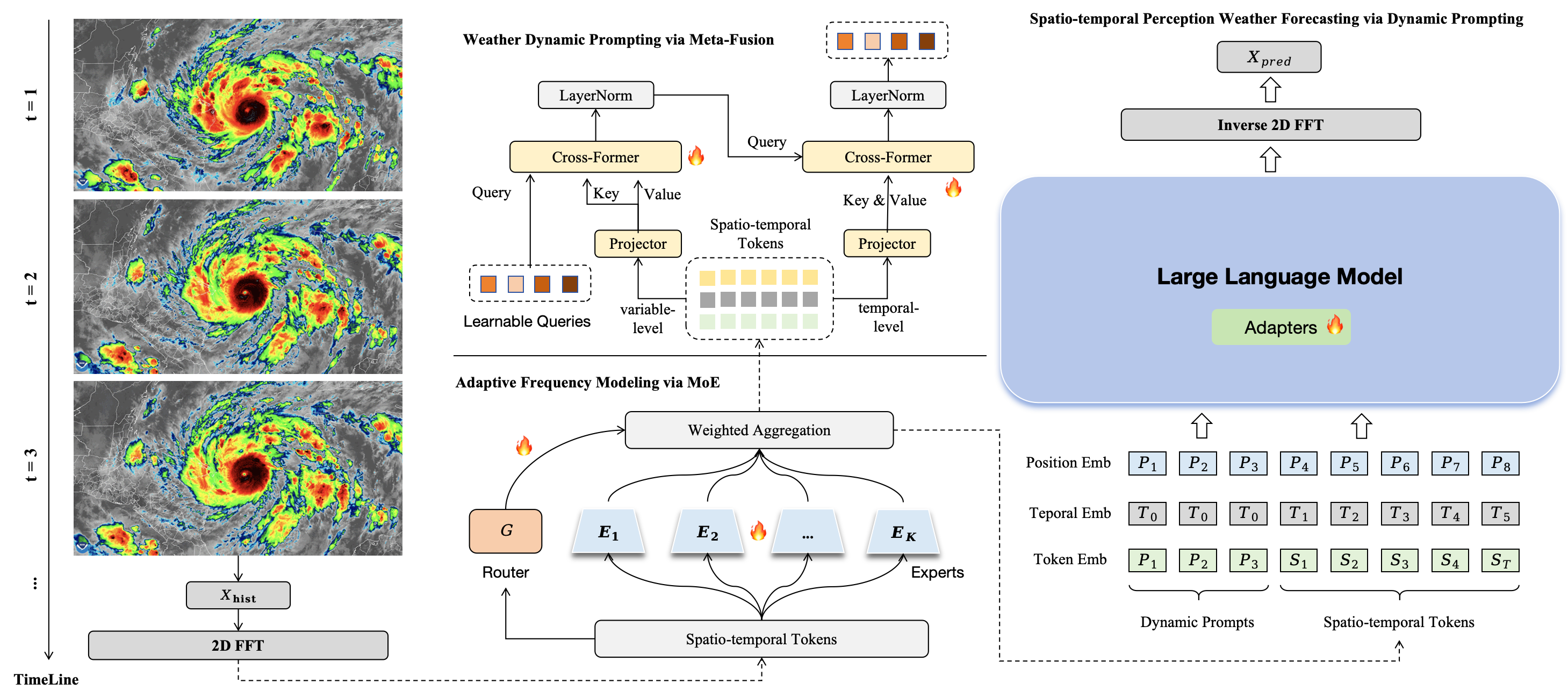}
\caption{Overall framework of the proposed ClimateLLM. (a) The two-dimensional time-series weather data \( X_{\text{hist}} \) is transformed into the frequency domain via 2D FFT. (b) A Mixture-of-Experts approach adaptively learns different frequency components. (c) Learnable prompts at the weather variable and temporal levels perform cross-attention for meta fusion. (d) The prompts and frequency domain tokens are fed into an LLM to capture spatiotemporal patterns, yielding predictions \( X_{\text{pred}} \).}
  \label{M1}
\end{figure*}

In this section, we mainly introduce \textbf{ClimateLLM} (Figure~\ref{M1}), a framework that integrates frequency-domain representation, dynamic prompting, and a Generative Pre-trained Transformer (GPT) backbone for weather forecasting. 

\subsection{Representation Learning via Frequency Mixture-of-Experts (MoE)}
\label{subsec:fmoe}

Accurate weather forecasting requires learning both spatial and temporal relationships in complex meteorological systems. 
For example, reliable temperature or precipitation predictions must capture large-scale circulation patterns (e.g., global wind jets, synoptic fronts) as well as local, fast-changing phenomena (e.g., convection, thunderstorms). 
Extreme weather events—such as severe convective storms, tropical cyclones, or atmospheric rivers—amplify this challenge: they involve strong nonlinear interactions, evolve rapidly, and often have distinct frequency signatures. 
recent studies have investigated deep learning approaches for predicting thunderstorm severity using remote sensing weather data\cite{9769919}, demonstrating the potential of advanced neural architectures to capture complex meteorological signals.
We found that while patch-based CNN or GNN approaches are intuitive for spatial feature extraction, they offer only limited gains for these highly localized events, especially when integrated with LLMs that excel at sequence-based reasoning but do not inherently resolve spatial structures.


Motivated by recent progress in Fourier-based neural operators (FNOs), we adopt a frequency-domain view to address these issues in extreme weather forecasting. 
Rather than subdividing input grids into patches for a CNN or creating graph structures for a GNN, we apply a two-dimensional Fourier Transform (2D FFT) to each spatial slice. 
This converts the data from the spatial domain into the frequency domain, revealing both low-frequency (broad-scale) and high-frequency (fine-scale) details without explicit local convolutions or adjacency matrices. 
For extreme weather events, frequency-domain modeling can uncover wavenumber patterns associated with severe storms or other wave-like processes—patterns that are often harder to identify in the raw spatial domain.


Still, not every frequency component is equally important for prediction. 
Most Fourier-based methods process these components uniformly, ignoring differences between low- and high-frequency bands. 
This oversimplifies the modeling of extreme phenomena. In our approach, a frequency-based MoE module adaptively allocates different expert networks to different segments of the frequency spectrum. 
We then use an LLM as the primary sequence learner, leveraging its capability for pattern extraction over extended temporal contexts. By combining frequency-domain representations with the LLM’s temporal insights, our framework addresses both quick local disturbances and broader-scale dependencies. The frequency pathway injects domain-specific structure, while the LLM refines long-range temporal patterns.

\subsubsection{Normalization and Frequency-Domain Representation}

Raw climate data often span different scales across variables. To manage this, each variable is normalized by subtracting its mean and dividing by its standard deviation over the historical period. Formally, at time \(t\):

\begin{equation}
\hat{X}(t)[v, m, n] = \frac{X(t)[v, m, n] - \mu(v, t)}{\sigma(v, t) + \epsilon},
\end{equation}
where \(\mu(v, t)\) and \(\sigma(v, t)\) are the mean and standard deviation of variable \(v\) at time \(t\), computed over the historical sequence. Once normalized, the data is transformed into the frequency domain using the 2D Fast Fourier Transform (2D FFT):
\begin{equation}
S(t) = \mathcal{F}\bigl(\hat{X}(t)\bigr),
\end{equation}
where the 2D FFT is:
\begin{equation}
S(t)[v, k_m, k_n] 
= \sum_{m=1}^{M}\sum_{n=1}^{N}\hat{X}(t)[v, m, n]\, e^{-2\pi i \bigl(\frac{k_m m}{M} + \frac{k_n n}{N}\bigr)}.
\end{equation}
Indices \(k_m\) and \(k_n\) represent frequencies along the two spatial dimensions. Lower frequencies capture global structures, while higher frequencies represent fine-scale variations. The result is a complex-valued representation:
\begin{equation}
S(t)[v, k_m, k_n] = \mathcal{R}(t)[v, k_m, k_n] + i\, \mathcal{I}(t)[v, k_m, k_n],
\end{equation}
where \(\mathcal{R}(t)\) and \(\mathcal{I}(t)\) are the real and imaginary parts, respectively.



\subsubsection{Mixture of Experts for Adaptive Frequency Modeling}

Distinct spatial patterns arise at different frequencies. To model them effectively, we introduce a MoE module that adaptively routes each frequency component to the most suitable sub-network. Let

\begin{equation}
Z(t) = g\bigl(S(t)\bigr),
\end{equation}
where \(g(\cdot)\) is a learnable transformation. The MoE includes \(E\) experts \(\{f_e(\cdot)\}_{e=1}^E\), each specializing in part of the frequency domain:

\begin{equation}
\tilde{S}(t) 
= \sum_{e=1}^{E} G_e\bigl(S(t)\bigr)\, f_e\bigl(Z(t)\bigr).
\end{equation}
Here, \(G_e(\cdot)\) is a gating function that assigns a weight to each expert’s output, ensuring a soft selection process. This allows the model to handle high- and low-frequency patterns together.

\subsubsection{LLM Integration for Temporal Dependencies}
After the MoE layer, we obtain feature representations that combine temporal hidden representations with frequency-domain information transformed from the spatial domain. Since weather variations are influenced not only by spatial factors but also by temporal evolution patterns, capturing the underlying temporal dependencies is crucial. Generative Pre-trained Transformers (GPTs) have demonstrated exceptional capabilities in sequence representation and pattern extraction and have been widely applied to time series forecasting tasks. Inspired by this, we further incorporate GPT to capture the temporal evolution patterns of spatial-frequency representations. Specifically, we treat the transformed spectral representation \(\tilde{S}(t)\) at each time step \(t\) as a token and leverage the self-attention mechanism to model the temporal dependencies between these tokens, denoted as $\tilde{H}=GPT(\tilde{S})$. This provides a deeper understanding of the temporal evolution, beyond what simpler CNN- or GNN-based structures might glean.

\subsubsection{Inverse Fourier Transform for Spatial Reconstruction}
After processing in the frequency domain, we apply the inverse 2D FFT (iFFT) to reconstruct the spatial representation:
\begin{equation}
\tilde{X}_{\text{pred}}(t) = \mathcal{F}^{-1}(\tilde{H}(t))
\end{equation}
where the inverse transformation is computed as:
\begin{align}
\mathcal{F}^{-1}(\tilde{H}(t))[v, m, n] &= \frac{1}{MN} 
\sum_{k_m=1}^{M} \sum_{k_n=1}^{N} \tilde{H}(t)[v, k_m, k_n] \notag \times e^{2\pi i (k_m m / M + k_n n / N)}
\end{align}
The de-normalization operator \( R_{de} \) acts on the inverse-transformed representation to obtain the final predicted climate state.:
\begin{equation}
X_{\text{pred}}(t) = R_{de}\left(\tilde{X}_{\text{pred}}(t)\right)
\end{equation}
The complete algorithm workflow is described in Algorithm~\ref{alg:fft_pipeline}. 

\subsubsection{Proposition}
We further have the following proposition (for the full proof, please refer to Appendix~\ref{app:fnoprove}):

\textbf{Proposition 1 (Equivalence of Time-Domain Forecasting and Frequency-Domain Forecasting for 2D FNO)}

\textit{Assume \( \{ (x_0, y_0), (x_1, y_1), \dots, (x_{N-1}, y_{N-1}) \} \) is the input sequence in the time domain, and \( \{ (\hat{x}_0, \hat{y}_0), (\hat{x}_1, \hat{y}_1), \dots, (\hat{x}_{N}, \hat{y}_{N}) \} \) is the predicted output sequence of the frequency model. The predicted value \( (\hat{x}_N, \hat{y}_N) \) is obtained by transforming from the frequency domain to the time domain at timestamp \( N \).}

\subsection{Weather Dynamic Prompting via Meta-Fusion}
\label{sec:dynamic_prompting}

Prompting has emerged as a technique for providing feature patterns or guidance tokens that steer LLMs toward more effective sequence forecasting. For instance, TimeLLM~\cite{jin2023time} combines domain knowledge and temporal statistics into prompt tokens to better inform the underlying LLM on where to focus. Despite these advances, many existing prompt designs rely on hard-coded information and thus struggle to capture dynamic temporal patterns. Moreover, unlike purely temporal tasks, weather forecasting also demands strong spatial modeling. Meteorological variables often propagate across space (e.g., storm fronts spreading geographically), while different variables (such as temperature and pressure) exhibit intricate correlations governed by atmospheric physics. 

To address these issues, we propose a weather dynamic prompting via meta-fusion strategy. Our design aims to capture the evolving temporal patterns while simultaneously bridging cross-variable, spatiotemporal information. Rather than directly encoding domain priors in rigid ways, we introduce learnable tokens into the LLM pipeline as queries in a cross-attention mechanism. This two-step “meta-fusion” not only diverges from traditional hard-encoding approaches, but also extends beyond simple concatenation or pooling along time axes. By doing so, it simultaneously captures crucial temporal patterns while acting as a powerful “bridge” to harness global weather information in both time and variable dimensions.

Formally, let 
\(\mathbf{P} \in \mathbb{R}^{K \times d}\)
denote the learnable prompt tokens, where \(K\) is the number of prompt tokens and \(d\) is the hidden dimension. Suppose we have a representation 
\(\mathbf{\tilde{S}} \in \mathbb{R}^{C \times L \times d}\)
obtained from the MoE block, where \(C\) is the number of weather variables and \(L\) is the length of the temporal sequence. We first aggregate along the variable dimension to obtain a purely temporal representation
\(\mathbf{\tilde{S}}_{t} \in \mathbb{R}^{L \times d}\). Then we perform cross-attention by taking the learnable tokens \(\mathbf{P}\) as queries and \(\mathbf{\tilde{S}}_{t}\) as both keys and values:
\begin{equation}
\mathbf{P}' \;=\;
\mathrm{LayerNorm}\Bigl(
    \mathrm{CrossAttn}\bigl(\mathbf{P},\, \mathbf{\tilde{S}}_{t},\, \mathbf{\tilde{S}}_{t}\bigr)
    \;+\;
    \mathbf{P}
\Bigr),
\end{equation}
where $\mathrm{CrossAttn}$ denotes the cross-attention function. Next, we aggregate along the time dimension of \(\mathbf{S}\) to obtain a representation
\(\mathbf{\tilde{S}}_{c} \in \mathbb{R}^{C \times d}\)
that focuses on the variable-wise features (e.g., aggregated temporal patterns for each variable). We again use \(\mathbf{P}\) as queries, but this time attend over \(\mathbf{\tilde{S}}_{c}\):
\begin{equation}
\tilde{\mathbf{P}} \;=\;
\mathrm{LayerNorm}\Bigl(
    \mathrm{CrossAttn}\bigl(\mathbf{P}',\, \mathbf{\tilde{S}}_{c},\, \mathbf{\tilde{S}}_{c}\bigr)
    \;+\;
    \mathbf{P}'
\Bigr).
\end{equation}
Here, the two cross-attention steps exploit the prompt tokens both as flexible probes of temporal dynamics and as a fusion bridge across different weather variables.

\subsection{Generative Pre-trained Transformer Backbone}
\label{sec:transformer_backbone}

The Generative Pre-trained Transformer (GPT) architecture, which underpins modern LLMs, leverages self-attention mechanisms to model long-range dependencies in sequential data. This makes LLMs particularly well-suited for capturing complex temporal patterns and dynamics. In temporal modeling applications, LLMs offer several key advantages in capturing both short-term fluctuations and long-term trends. To further enhance the temporal representation, we integrate an additional time-series encoding that complements the standard transformer positional encodings. Specifically, our framework processes the input data through a two-dimensional Fast Fourier Transform (2DFFT) and a Mixture-of-Experts (MoE) module to extract salient features. This yields a set of MoE representations, denoted as $\tilde{S}$, and weather prompts, denoted as $\tilde{P}$. We then concatenate these outputs and feed them into the LLM as follows:
\begin{equation}
  \mathbf{\tilde{H}} 
  \;=\; 
  \mathrm{GPT}\!\Bigl(
    \mathrm{concat}\bigl[\widetilde{\mathbf{P}},\, 
    \widetilde{\mathbf{S}}_{1}, \widetilde{\mathbf{S}}_{2}, \ldots, \widetilde{\mathbf{S}}_{T}\bigr]
  \Bigr),
\end{equation}
In line with recent developments in LLM-based temporal foundation models \cite{pan2024s,cao2023tempo}, our approach adopts the GPT-2 architecture as the backbone. GPT-2 is renowned for its scalable transformer design, efficient self-attention mechanism, and robust performance on sequence modeling tasks.

\subsection{Latitude-Weighted Training and Optimization}
\label{subsec:train_opt}
In this paper, we employ the latitude-weighted Root Mean Square Error (RMSE) as the optimization objective instead of the conventional RMSE. Traditional RMSE treats all spatial grid points equally, assuming a uniform distribution of errors across the dataset. However, in global weather modeling, the Earth is a sphere, and data points at higher latitudes (closer to the poles) are disproportionately represented in gridded datasets due to the convergence of meridians. This introduces a latitude bias, where errors in high-latitude regions can disproportionately influence the overall RMSE, leading to an inaccurate assessment of model performance.

To mitigate this issue, we adopt latitude-weighted RMSE, where each grid point is weighted according to its latitude. The weight is defined as:
\begin{equation}
    \alpha(m) = \frac{\cos(m)}{\sum_{m'} \cos(m')}
\end{equation}
where \( m \) represents the latitude index. This weighting scheme ensures that errors in lower latitudes, which cover larger surface areas, contribute proportionally more to the loss function, aligning the optimization objective with the actual physical characteristics of the Earth’s surface.

The latitude-weighted RMSE is formulated as:
\begin{equation}
    Loss = \sqrt{\frac{1}{MN}\sum_{m=1}^{M}\sum_{n=1}^{N}\alpha(m)(\mathbf{X}_\text{pred}(m,n) - \mathbf{X}_\text{true}(m,n))^2}
\end{equation}
\( X_{\text{pred}} \) represents the prediction result, which is obtained by applying inverse 2DFFT and de-normalization to the representation \( \tilde{H} \).

\section{EXPERIMENTS}

In this section, we conduct extensive experiments to answer the following questions:
\begin{itemize}
\item \textbf{RQ1} How does our ClimateLLM model perform compared to the state-of-the-art methods?
\item \textbf{RQ2} To what extent does our proposed model improve training and inference efficiency compared to existing methods?
\item \textbf{RQ3} How does our method perform as a foundation model in zero-shot and few-shot prediction?
\item \textbf{RQ4} How do the key components and modules of the model affect its performance?
\item \textbf{RQ5} What impact do the model's hyperparameter settings have on its performance?
\item \textbf{RQ6} How does the model perform in real-world extreme weather prediction cases?
\end{itemize}

\subsection{Experimental Settings}

\subsubsection{Datasets}

\begin{table}
\centering
  \caption{ Variables of the ERA5 datasets.}
  \label{tab:datasets Statistics}
  \begin{tabular}{cccc}
    \toprule
    Variable name & Abbrev. &  ECMWF ID & Levels \\
    \midrule
    2 meter temperature & t2m & 167 & - \\
    10 meter U wind component & u10 & 165 & - \\
    Geopotential& z & 129 & 500\\
    Temperature & t & 130 & 850\\
  \bottomrule
\end{tabular}
\label{dataset}
\end{table}
In this study, we utilize the ERA5 reanalysis dataset \cite{hersbach2020era5,rasp2024weatherbench}, which is the fifth generation ECMWF atmospheric reanalysis of the global climate. We specifically employ the 5.625-degree resolution version (64×32 grid points) of ERA5 from 2006 to 2018, which provides comprehensive atmospheric data at various pressure levels. Four key atmospheric variables described in the Table \ref{dataset} are selected for our analysis.

\subsubsection{Evaluation Metrics}
In this paper, we focus mainly on the precision of the prediction of weather variables. Following related work \cite{rasp2024weatherbench}, there are two metrics to evaluate the prediction accuracy, namely Root mean squared error (RMSE) and Anomaly correlation coefficient (ACC). Due to the varying grid cell areas in the equiangular latitude-longitude grid system (where polar cells are smaller than equatorial cells), we apply area-weighted metrics across grid points to prevent polar bias. The detailed definitions of the latitude-weighted RMSE and ACC can be found in Appendix~\ref{app:metrics}.

\subsubsection{Baselines}
To fairly and effectively evaluate the performance of our model, we compared it with various state-of-the-art methods under the same experimental settings:
\begin{itemize}
\item \textbf{NODE} \cite{chen2019neuralordinarydifferentialequations}: Neural Ordinary Differential Equations (NODE) model is a continuous-depth neural network model and uses differential equation solvers to compute outputs by parameterizing the derivatives of hidden states.
\item \textbf{FourCastNet} \cite{pathak2022fourcastnet}: 
FourCastNet is a deep learning model developed for global weather forecasting that uses the Vision Transformer (ViT) and Fourier Neural Operator (FNO) architecture for weather prediction.
\item \textbf{ClimaX} \cite{nguyen2023climax}: 
ClimaX is a foundation model using self-supervised learning for weather and climate science that uses a transformer-based architecture to handle multiple types of Earth system data.
\item \textbf{ClimODE} \cite{verma2024climode}: ClimODE implements weather prediction as a physics-informed neural ODE based on the principle of advection. It models weather as a continuous-time transport process through a hybrid neural network combining local convolutions and global attention.

\end{itemize}

\subsubsection{Parameter Settings}
We split the ERA5 dataset (2006-2018) into training set (2006-2015), validation set (2016) and test set (2017-2018). The hyperparameter of the baseline models are set according to the corresponding optimal parameters. The batch size is set as 64 and the learning rate is set as 1e-3. All models use the Adam optimizer \cite{kingma2017adammethodstochasticoptimization} for parameter updates.

\subsection{Overall Performance (RQ1)}

\subsubsection{ACC} 
\begin{table}
\tiny
\renewcommand{\arraystretch}{1.3}
  \caption{Performance comparison of different models on weather forecasting tasks. The table shows ACC metrics across different variables and lead times.}
  \label{tab:model_performance}
  \setlength{\tabcolsep}{0.8pt}
  \begin{tabular}{c cccc cccc cccc}
    \toprule
    \multirow{2}{*}{Model} & \multicolumn{4}{c}{\textbf{z}} & \multicolumn{4}{c}{\textbf{t}} & \multicolumn{4}{c}{\textbf{t2m}}  \\
    \cmidrule(lr){2-5} \cmidrule(lr){6-9} \cmidrule(lr){10-13}
    & 6h & 12h & 18h & 24h & 6h & 12h & 18h & 24h & 6h & 12h & 18h & 24h \\
    \midrule
    NODE & 0.96 & 0.88 & 0.79 & 0.70 & 0.94 & 0.85 & 0.77 & 0.72 & 0.82 & 0.68 & 0.69 & 0.79 \\
    ClimaX & 0.97 (+1\%) & 0.96 (+9\%) & 0.95 (+20\%) & 0.93 (+33\%) & 0.94 (+0\%) & 0.93 (+9\%) & 0.92 (+20\%) & 0.90 (+25\%) & 0.92 (+12\%) & 0.90 (+32\%) & 0.88 (+28\%) & 0.89 (+13\%) \\
    FCN & \underline{0.99} (+3\%) & \underline{0.99} (+13\%) & \underline{0.99} (+25\%) & \underline{0.99} (+41\%) & \underline{0.99} (+5\%) & \underline{0.99} (+17\%) & \underline{0.99} (+29\%) & \textbf{0.99} (+38\%) & \underline{0.99} (+21\%) & \underline{0.99} (+46\%) & \underline{0.99} (+44\%) & \underline{0.99} (+25\%) \\
    ClimODE & 0.99 (+3\%) & 0.99 (+13\%) & 0.98 (+24\%) & 0.98 (+40\%) & 0.97 (+3\%) & 0.96 (+13\%) & 0.96 (+25\%) & 0.95 (+32\%) & 0.97 (+18\%) & 0.96 (+41\%) & 0.96 (+39\%) & 0.96 (+22\%) \\
    \hline
    ClimateLLM & \textbf{1.00} (+4\%) & \textbf{1.00} (+14\%) & \textbf{0.99} (+25\%) & \textbf{0.99} (+41\%) & \textbf{1.00} (+6\%) & \textbf{0.99} (+17\%) & \textbf{0.99} (+29\%) & \underline{0.98} (+36\%) & \textbf{1.00} (+22\%) & \textbf{1.00} (+47\%) & \textbf{0.99} (+44\%) & \textbf{0.99} (+25\%) \\
    \bottomrule
  \end{tabular}
\end{table}

To verify the effectiveness of our proposed model, we conducted comprehensive experiments on the ERA5 dataset with different prediction horizons, and the ACC forecasts results for those four variables in the next 24 hours are shown in Table~\ref{tab:model_performance}. Analyzing the experimental results, we have the following observations:

\begin{itemize}
\item Superior Performance: ClimateLLM consistently achieves higher ACC scores across all variables, with exceptional performance in t2m predictions maintaining 0.98-1.00 ACC values.
\item Temporal Robustness: The model exhibits minimal performance degradation over extended forecast horizons (6h-24h), significantly outperforming both traditional and deep learning baselines.
\item Exceptional Anomaly Prediction Capability: ClimateLLM's consistently high ACC scores (0.98-1.00) across variables, particularly in t2m predictions through 24h lead time, demonstrates not only superior mathematical accuracy but also remarkable meteorological significance - the model exhibits profound understanding of weather system dynamics and anomaly patterns, enabling accurate prediction of extreme weather events and reliable medium-range forecasts. 

\end{itemize}

\subsubsection{RMSE} 
\begin{table}[h]
\centering
\setlength{\tabcolsep}{4pt}
\caption{Comparison of different models' RMSE metrics variables at lead times of 6 hours.}
\begin{tabular}{ccccccc}
\hline
\multirow{2}{*}{Variable} & \multicolumn{5}{c}{RMSE($\downarrow$)} \\
\cline{2-6}
  &NODE & ClimaX & FCN & ClimODE & ClimateLLM \\
\hline
z  & 300.64 & 247.5 & 149.4 & \textbf{112.3} & \underline{143.2}\\
\hline
t  & 1.82 & 1.64 & \underline{1.18} & 1.19 & \textbf{1.04}\\
\hline
t2m  & 2.72 & 2.02 & 1.28 & \underline{1.27} & \textbf{1.02 }\\
\hline
u10  & 2.3 & 1.58 & \underline{1.47} & 1.48 & \textbf{1.46}\\

\hline
\end{tabular}
\label{rmse_results}
\end{table}
Based on the RMSE metrics comparison at 6-hour lead times at Table \ref{rmse_results}, ClimateLLM demonstrates superior performance across multiple variables compared to other models at short-term forecasting task. Specifically, for the temperature (t), ClimateLLM's RMSE of \textbf{1.04} shows a 42.9\% reduction compared to NODE (1.82) and a 11.9\% improvement over FCN (1.18). Similarly, in 2 meter temperature predictions (t2m), ClimateLLM exhibits an RMSE of \textbf{1.02}, marking a 19.7\% improvement over ClimODE (1.27). The results demonstrate ClimateLLM's exceptional performance in short-term temperature prediction tasks at 6-hour lead times.

\subsubsection{Long-term weather forecasting task}
\begin{table}
\centering
\caption{Longer lead time predictions.}
\label{tab:longer_lead_time}
\begin{tabular}{ccccc}
\hline
Variable & Lead-Time & \multicolumn{3}{c}{ACC($\uparrow$)} \\
\cline{3-5}
& (hours)& ClimaX & ClimODE & ClimateLLM\\
\hline
\multirow{2}{*}{z} & 72 & 0.73 & \underline{0.88} & \textbf{0.94}\\
& 144 & 0.58 & \underline{0.61} & \textbf{0.89}\\
\hline
\multirow{2}{*}{t} & 72 & 0.76 & \underline{0.85} & \textbf{0.95}\\
& 144 & 0.69 & \underline{0.77} & \textbf{0.94}\\
\hline
\multirow{2}{*}{t2m} & 72 & 0.83 & \underline{0.85} & \textbf{0.98}\\
& 144 & \underline{0.83} & 0.79 & \textbf{0.96}\\
\hline
\multirow{2}{*}{u10} & 72 & 0.45 & \textbf{0.66} & \underline{0.61}\\
& 144 & 0.30 & \underline{0.35} & \textbf{0.52}\\
\hline
\end{tabular}
\end{table}
The results demonstrate ClimateLLM's capabilities in long-term weather prediction tasks. In Table \ref{tab:longer_lead_time}, at extended lead times of 72 and 144 hours, ClimateLLM consistently outperforms baseline models across all variables. Particularly noteworthy is its performance in temperature forecasting, where it achieves exceptional ACC scores at 72 and 144 hours, showing improvements of 11.8\% and 22.1\% over ClimODE. For 2-meter temperature (t2m), ClimateLLM demonstrates even stronger performance, outperforming ClimODE by 15.3\% and 21.5\%. These substantial improvements underscore ClimateLLM's robust predictive capabilities in capturing long-term temperature dynamics.

\subsection{Model efficiency (RQ2)}
\begin{table}[h]
    \centering
        \caption{Efficiency Performance Comparison.}
    \begin{tabular}{lccc}
        \hline
        Model & Training & GPU Memory & Total Training\\
        & Time (s/epoch) & Usage (MB) & Time (hours)\\
        \hline
        ClimODE & 212.76 & 34,900 & 17.6\\
        ClimateLLM & \textbf{26.65} & \textbf{2,564} & \textbf{0.22}\\
        \hline
    \end{tabular}

    \label{tab:model_comparison}
\end{table}
In terms of model efficiency, ClimateLLM demonstrates dramatic improvements over the baseline ClimODE across all computational metrics. As shown in Table \ref{tab:model_comparison}, the training time per epoch is reduced from 212.76 seconds to just \textbf{26.65} seconds, representing an impressive \textbf{87.5\%} reduction. More remarkably, ClimateLLM achieves a substantial decrease in GPU memory consumption, requiring only \textbf{2,564} MB compared to ClimODE's 34,900 MB - a remarkable \textbf{92.7\%} reduction in memory usage. Perhaps most significantly, the total training time is reduced from 17.6 hours to merely \textbf{0.22} hours, marking a \textbf{98.7\%} improvement in overall training efficiency. These substantial enhancements in computational efficiency demonstrate ClimateLLM's superior resource utilization while maintaining its strong predictive performance.

\subsection{Zero-Shot and Few-Shot Forecasting (RQ3)}

\begin{figure}[h]
  \centering
  \includegraphics[width=0.8\linewidth]{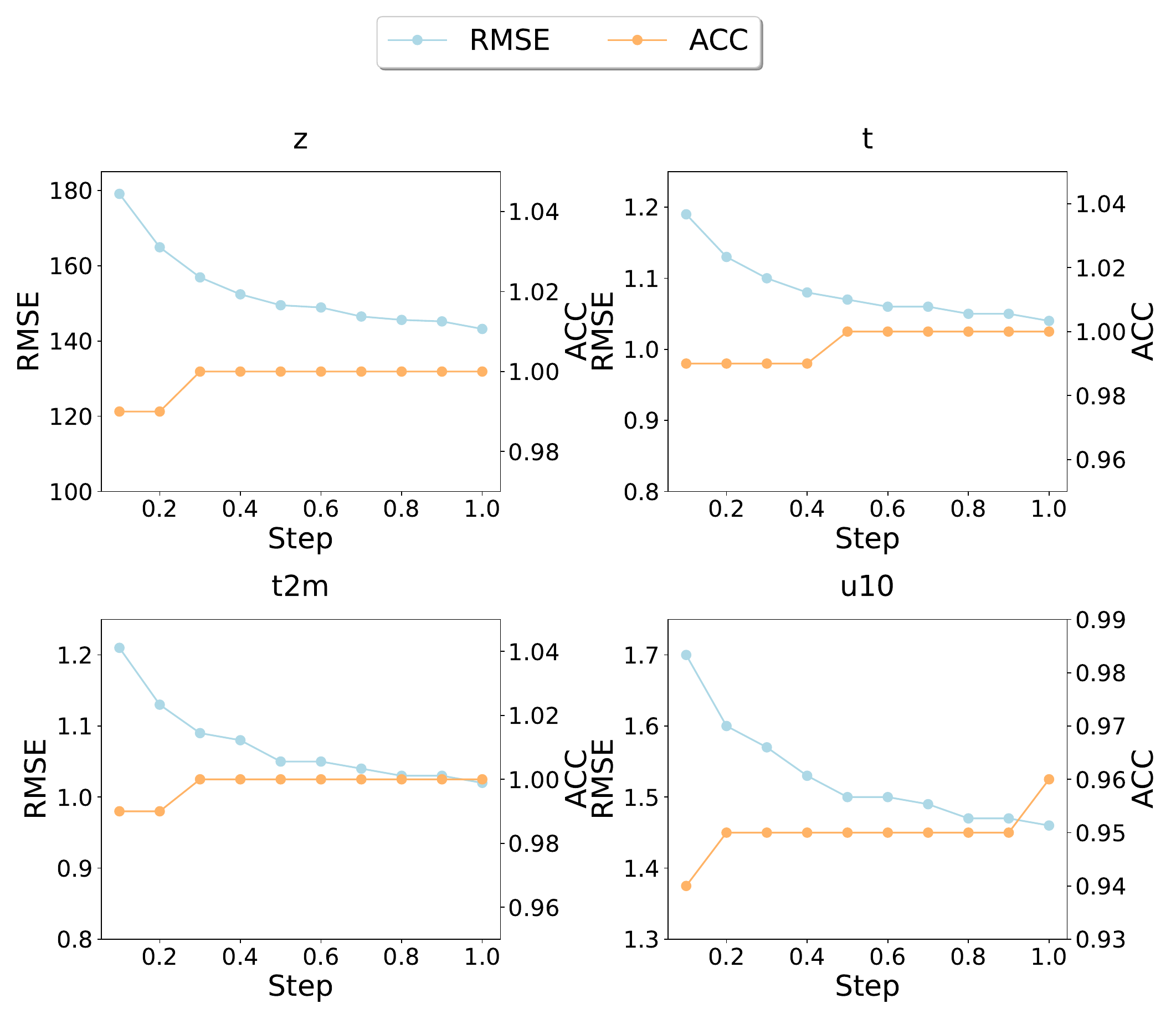}
  \caption{Few-shot Forecasting Results, with training sample scale ranging from 10\% to 100\%.}
  \label{few_shot}
\end{figure}

\begin{table}[h]
    \centering
    \begin{tabular}{ccc}
        \hline
        \textbf{Prediction Task} & \textbf{Metric} & \textbf{Value} \\
        \hline
        \multirow{2}{*}{$t$ $\rightarrow$ $t2m$} & RMSE & 2.07 \\
        & ACC & 0.99 \\
        \hline
        \multirow{2}{*}{$t2m$ $\rightarrow$ $t$} & RMSE & 1.28 \\
        & ACC & 0.99 \\
        \hline
    \end{tabular}
    \caption{Zero-shot Forecasting Results. Left of the arrow $\rightarrow$ training samples, right $\rightarrow$ test samples.}
    \label{tab:zeroshot_results}
\end{table}
To validate the generalization ability of our method as a foundation model, we evaluated its performance under both zero-shot and few-shot forecasting settings. As shown in Table \ref{tab:zeroshot_results} for the experiments on \(t\) and \(t2m\), our approach demonstrates strong zero-shot prediction capability. The ACC for \(t2m\) reaches 0.99, exceeding the full-shot ClimODE’s 0.96. Similarly, the ACC for \(t\) is 0.99, outperforming ClimODE’s 0.97, while its RMSE of 1.28 represents only a slight degradation compared to ClimODE’s 1.19. For the few-shot experiments (illustrated in Figure~\ref{few_shot}), the proportion of training samples is incrementally increased from 0.1 to 1.0. Notably, when only 20\% of the training data is used, the ACC for all three variables (\(z\), \(t\), and \(t2m\)) reaches 0.99—surpassing the performance of ClimODE and ClimaX models trained on the full dataset. Similarly, the RMSE, which directly reflects overall prediction accuracy, significantly outperforms the baseline methods even with just 20\% of the training samples. These experimental results robustly validate the zero-shot and few-shot capabilities of our ClimateLLM as a foundation model, a success largely attributable to our design of a frequency-aware LLM with dynamic prompting.

\subsection{Ablation Experimental Study (RQ4)}

\begin{table}

  \caption{Ablation study on key components of ClimateLLM.}

  \label{tab:different module preformance}
  \begin{tabular}{lcccccccc}
    \toprule
    \multirow{2}{*}{Model} & \multicolumn{2}{c}{\textbf{z}} & \multicolumn{2}{c}{\textbf{t}} & \multicolumn{2}{c}{\textbf{t2m}} & \multicolumn{2}{c}{\textbf{u10}}\\ 
    \cmidrule(lr){2-3}  \cmidrule(lr){4-5}\cmidrule(lr){6-7}\cmidrule(lr){8-9}
    & RMSE$\downarrow$ & ACC$\uparrow$ & RMSE$\downarrow$ & ACC$\uparrow$ & RMSE$\downarrow$ & ACC$\uparrow$ & RMSE$\downarrow$ & ACC$\uparrow$\\
    \midrule
    ClimateLLM & \textbf{143.2} & \textbf{1.00}  & \textbf{1.04} & \textbf{1.00} & \textbf{1.02} & \textbf{1.00}& \textbf{1.46} & \textbf{0.96} \\
    \hline
    w/o FFT & 153.5 & 0.97 & 1.12 & 0.97 & 1.23 & 0.96&1.61&0.93\\
    \hline
    w/o Prompt& 145.8 & 0.99 & 1.07& 0.99 & 1.05& 0.99& 1.49& 0.95\\
    \hline
    w/o MOE & 149.2 & 0.98 & 1.09 & 0.98 & 1.15 & 0.97 & 1.55 & 0.94\\
  \bottomrule
\end{tabular}
\end{table}

In order to study the influence of each module on the model effect, we consider conducting the following ablation experiments. (1) Without frequency domain transformation, (2) Without Prompt, (3) Without MOE. The experimental results are shown in Table ~\ref{tab:different module preformance}. We can observe the conclusions: After removing the FFT module, the model's performance decreased significantly (RMSE increased from 143.2 to 153.5, and ACC decreased from 1.00 to 0.97), indicating that FFT plays a more crucial role in the modeling process. Meanwhile, removing MOE and Prompt led to varying degrees of performance degradation, but with relatively smaller magnitudes (removing Prompt resulted in RMSE of 145.8 and ACC of 0.99; removing MOE resulted in RMSE of 149.2 and ACC of 0.98), suggesting that FFT is the key component affecting model performance, while MOE and Prompt serve supplementary optimization functions.

\subsection{Sensitive Analysis (RQ5)}
\begin{figure}[h]
  \centering
  \includegraphics[width=0.8\linewidth]{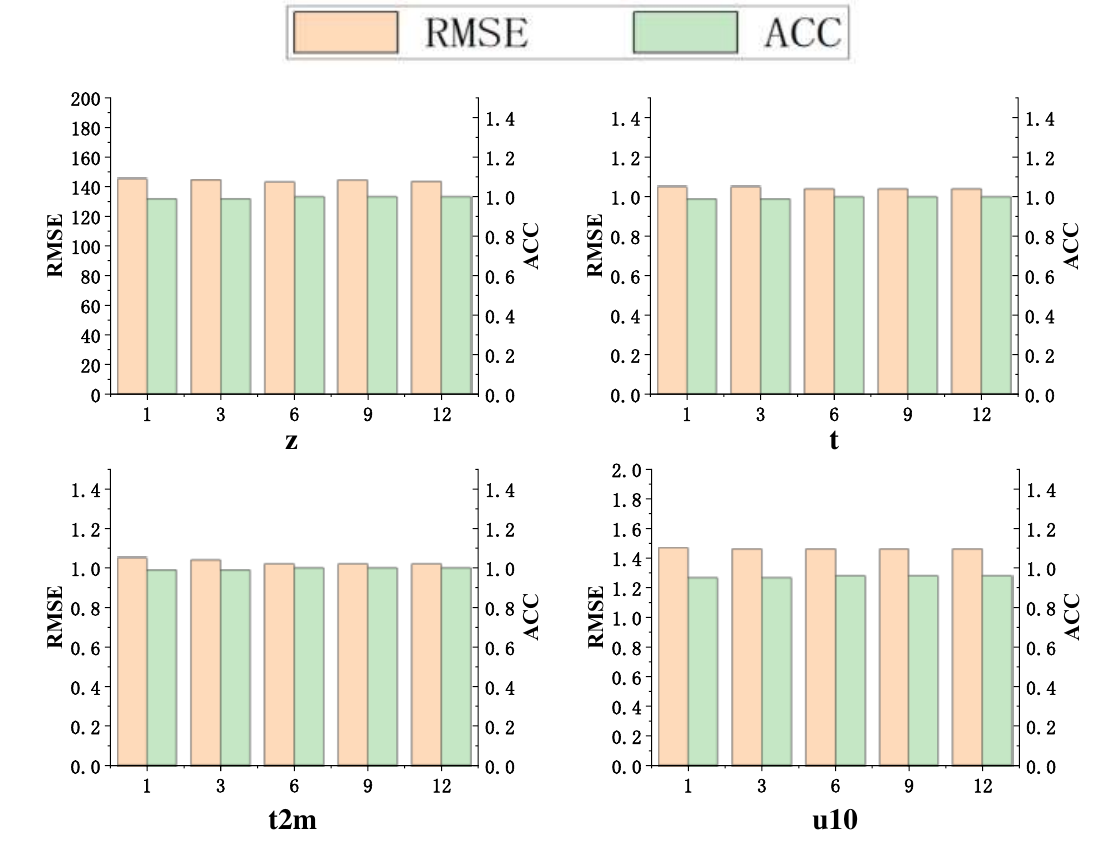}
  \caption{Sensitivity analysis of GPT's number of layers.}
  \label{s2}
\end{figure}

The generative pre-trained transformer serves as the primary backbone of our ClimateLLM, and its parameter size often determines the model’s representation capability at different levels. Therefore, in this section, we mainly analyze the sensitivity of the number of GPT layers. As demonstrated in Figure \ref{s2}, our experimental results reveal that varying the number of GPT layers (1, 3, 6, 9, and 12) produced negligible differences in both RMSE and ACC metrics across variables, suggesting that our model demonstrates low sensitivity to the quantity of GPT layers.

\subsection{Extreme Weather Case Analysis (RQ6)}

\begin{figure}[h]
  \centering
  \includegraphics[width=\linewidth]{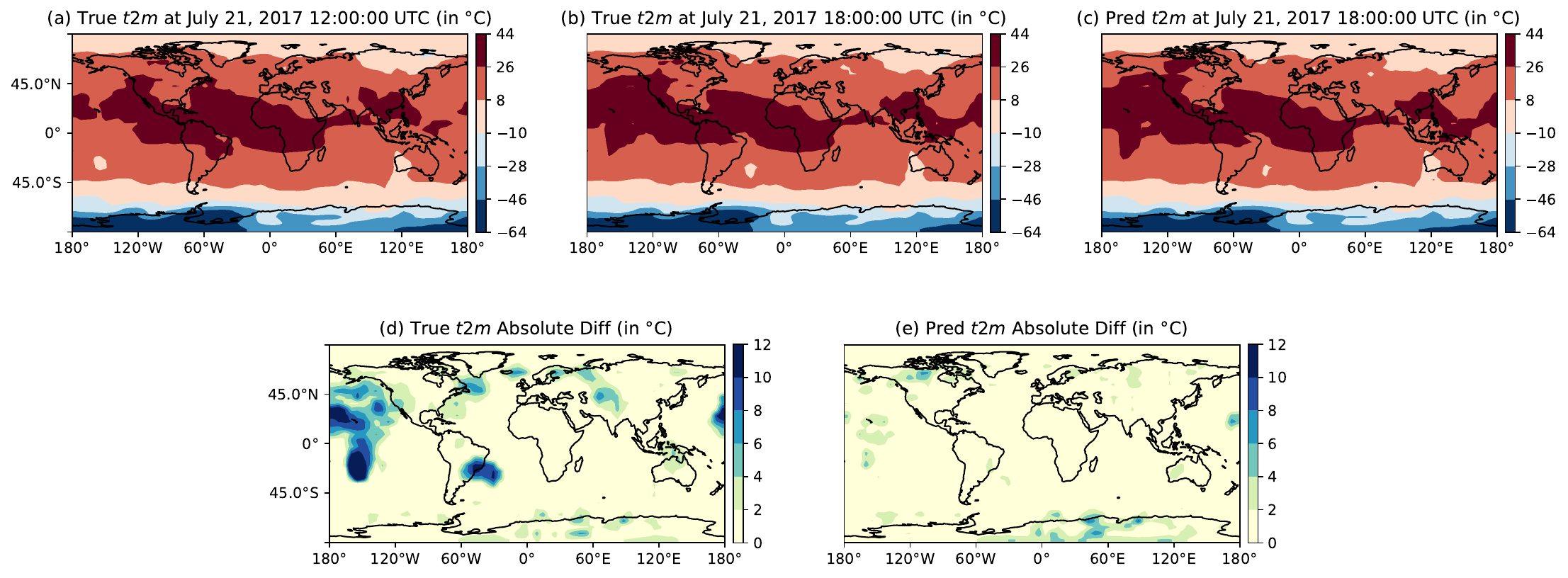}
  \caption{Case Study of variable $t2m$. (a) True vlaue at $t_0$ (b) True value at $t_1$ (c) ClimateLLM prediction results at $t_1$ (d) The difference between true value at $t_0$ and true value at $t_1$ (e) The difference between prediction result at $t_1$ and true value at $t_1$.}
  \label{t2m}
\end{figure}

In this section, we present a case study of extreme weather variation by examining the most significant temporal change in the $t2m$ variable from our test set (2017-2018). We selected a specific time step from July 21, 2017 12:00:00 UTC to July 21, 2017 18:00:00 UTC, which exhibited a temporal variation 11.5\% higher than the mean change, representing it's one of the most substantial fluctuations in our dataset. Given our model's exceptional performance in Anomaly Correlation Coefficient (ACC) prediction, it demonstrates remarkable capability in capturing such dramatic climate variable transitions. As illustrated in Figure \ref{t2m} (c), our model accurately predicted the temperature variation patterns over the Pacific Ocean adjacent to North America's western coast relative to the initial state shown in Figure \ref{t2m}(a). This temperature evolution is further corroborated by the differential map between the two time steps depicted in Figure \ref{t2m}(d). This case study validates our model's superior performance in predicting intense climate variations, demonstrating its exceptional capability in forecasting extreme weather events.


\section{Conclusion}
In this paper, we propose ClimateLLM, a weather forecasting foundation model based on frequency-domain perception. Our framework demonstrates that the combination of frequency-domain representation learning, dynamic prompting mechanisms, and pre-trained transformer models can effectively capture complex weather patterns while maintaining computational efficiency. Through extensive experiments across multiple meteorological variables and prediction horizons, we show that our approach achieves comparable or superior performance to state-of-the-art weather prediction systems, particularly in extreme weather events. The framework's ability to leverage pre-trained parameters while requiring minimal fine-tuning makes it particularly attractive for operational deployment. In the future, we plan to further explore several research directions. Firstly, we consider incorporating physics-informed neural networks to integrate prior weather physical knowledge into the model architecture, thereby helping the model more accurately and effectively capture both intra-variable and inter-variable weather dynamics. Secondly, we plan to introduce Tree-of-Thought-based reasoning algorithms, leveraging the powerful reasoning capabilities of LLMs to effectively capture temporal and spatial patterns of weather changes.


\clearpage


\medskip

\clearpage
{
\small
\bibliography{iclr2024_conference}
\bibliographystyle{iclr2024_conference}
}

\clearpage

\appendix

\section{Method}
The complete algorithm workflow is described in Algorithm~\ref{alg:fft_pipeline}.

\begin{algorithm}[H]
    \caption{2D FFT-based Climate State Processing Pipeline}
    \label{alg:fft_pipeline}
    \begin{algorithmic}[1]
        \Require Climate state $X(t) \in \mathbb{R}^{|\mathcal{V}| \times M \times N}$ at time $t$, historical sequence length $L$, number of experts $E$, and inverse transform function $\mathcal{F}^{-1}(\cdot)$.
        \Ensure Predicted climate state $X_{\text{pred}}(t) \in \mathbb{R}^{|\mathcal{V}| \times M \times N}$.

        \State \textbf{Data Normalization:}
        \State Compute mean and standard deviation:
        \begin{equation*}
            \mu(v, t) = \frac{1}{L M N} \sum_{l=t-L}^{t-1} \sum_{m=1}^{M} \sum_{n=1}^{N} X(l)[v, m, n]
        \end{equation*}
        \begin{equation*}
            \sigma^2(v, t) = \frac{1}{L M N} \sum_{l=t-L}^{t-1} \sum_{m=1}^{M} \sum_{n=1}^{N} (X(l)[v, m, n] - \mu(v, t))^2
        \end{equation*}
        \State Normalize data:
        \begin{equation*}
            \hat{X}(t)[v, m, n] = \frac{X(t)[v, m, n] - \mu(v, t)}{\sigma(v, t) + \epsilon}
        \end{equation*}

        \State \textbf{Apply 2D FFT:}
        \State Transform spatial data into the frequency domain:
        \begin{equation*}
            S(t) = \mathcal{F}(\hat{X}(t))
        \end{equation*}
        \begin{equation*}
            S(t)[v, k_m, k_n] = \sum_{m=1}^{M} \sum_{n=1}^{N} \hat{X}(t)[v, m, n] e^{-2\pi i (k_m m / M + k_n n / N)}
        \end{equation*}

        \State \textbf{Frequency Representation Learning using MoE:}
        \State Compute expert network outputs:
        \begin{equation*}
            Z(t) = g(S(t))
        \end{equation*}
        \begin{equation*}
            \tilde{S}(t) = \sum_{e=1}^{E} G_e(S(t)) f_e(Z(t))
        \end{equation*}
        where $G_e(S(t))$ is the gating function and $f_e(\cdot)$ represents the $e$-th expert.

        \State \textbf{Inverse 2D FFT for Spatial Reconstruction:}
        \State Convert processed frequency features back to spatial domain:
        \begin{equation*}
            \tilde{X}_{\text{pred}}(t) = \mathcal{F}^{-1}(\tilde{S}(t))
        \end{equation*}
        \begin{equation*}
            X_{\text{pred}}(t) = R_{de}\left(\tilde{X}_{\text{pred}}(t)\right)
        \end{equation*}
    \end{algorithmic}
\end{algorithm}

\section{Experimental Settings}
\subsection{Datasets}
We trained our model using the ERA5 datasets from WeatherBench2 \cite{rasp2024weatherbench}. WeatherBench 2 is a framework for evaluating and comparing data-driven and traditional numerical weather forecasting models. All data used in our experiments are availale at: \url{https://github.com/google-research/weatherbench2}
\label{app:datasets}
\subsection{Software and Hardware}
The model is implemented with PyTorch \cite{paszke2019pytorchimperativestylehighperformance} and the whole model training and inference is conducted on a single 80GB Nvidia A100 GPU.
\subsection{Metrics}
\label{app:metrics}

In this paper, we focus mainly on the precision of the prediction of weather variables. Following related work \cite{rasp2024weatherbench}, there are two metrics to evaluate the prediction accuracy, namely Root mean squared error (RMSE) and Anomaly correlation coefficient (ACC). Due to the varying grid cell areas in the equiangular latitude-longitude grid system (where polar cells are smaller than equatorial cells), we apply area-weighted metrics across grid points to prevent polar bias. The latitude weights $\alpha(m)$ are defined as:

\begin{equation}
    \alpha(m) = \frac{\cos(m)}{\sum_{m'} \cos(m')}
\end{equation}

where $m$ represents the latitude index of the grid point, and $L$ represents the latitude-dependent weighting factor used to account for the varying grid cell areas.

\begin{itemize}
    \item \textbf{Root mean squared error (RMSE)} The latitude-weighted RMSE for a forecast variable $v$ at forecast time-step $l$ is defined by the following equation, with the same latitude weighting factor given by Equation \ref{rmse},

\begin{equation}
\text{RMSE}(v) = \sqrt{\frac{1}{MN}\sum_{m=1}^{M}\sum_{n=1}^{N}\alpha(m)(\mathbf{X}_\text{pred}(m,n) - \mathbf{X}_\text{true}(m,n))^2}
\label{rmse}
\end{equation}

where $\mathbf{X}_\text{true/pred}(m,n)$ represents the value of predicted (/true) variable $v$ at the location denoted by the grid
co-ordinates $(m, n)$ at a forecast time-step.
    \item \textbf{Anomaly correlation coefficient (ACC)}
    The latitude weighted ACC for a forecast variable $v$ at forecast time-step $l$ is defined as follows:
\begin{equation}
\text{ACC}(v) = \frac{\sum_{m,n} L(m)\tilde{\mathbf{X}}_\text{pred}\tilde{\mathbf{X}}_\text{true}}{\sqrt{\sum_{m,n} L(m)\tilde{\mathbf{X}}_\text{pred}^2 \sum_{m,n} L(m) \tilde{\mathbf{X}}_\text{true}^2}}
\label{ACC}
\end{equation}

where $\tilde{\mathbf{X}}_\text{pred/true} = \mathbf{X}_\text{pred/true} - C$ represents the long-term-mean-subtracted value of predicted (/true) variable $v$. While $C = \frac{1}{N}\sum_{t}^{N}{\mathbf{X}_\text{true}}$ is the climatology mean of the history. For more detail, please refer to Appendix \ref{app:metrics}.
\end{itemize}

\section{Interpreting Model Predictions from Frequency Domain}
\label{app:fnoprove}
\textbf{Proposition 1 (Equivalence of Time-Domain Forecasting and Frequency-Domain Forecasting for 2D FNO)}

\textit{Assume \( \{ (x_0, y_0), (x_1, y_1), \dots, (x_{N-1}, y_{N-1}) \} \) is the input sequence in the time domain, and \( \{ (\hat{x}_0, \hat{y}_0), (\hat{x}_1, \hat{y}_1), \dots, (\hat{x}_{N}, \hat{y}_{N}) \} \) is the predicted output sequence of the frequency model. The predicted value \( (\hat{x}_N, \hat{y}_N) \) is obtained by transforming from the frequency domain to the time domain at timestamp \( N \).} 

\textit{Proof.} Assume \( \{ (x_0, y_0), (x_1, y_1), \dots, (x_{N-1}, y_{N-1}) \} \) is the input sequence in the time domain, and \( \{ (\hat{x}_0, \hat{y}_0), (\hat{x}_1, \hat{y}_1), \dots, (\hat{x}_{N}, \hat{y}_{N}) \} \) is the predicted output sequence of the frequency model. The predicted value \( (\hat{x}_N, \hat{y}_N) \) is obtained by transforming from the frequency domain to the time domain at timestamp \( N \). In this context, the prediction of the next frequency component \( F'(u,v) \) in the frequency domain allows for forecasting the next values in the time domain.

The 2D Discrete Fourier Transform (DFT) and its inverse (iDFT) are defined as:
\begin{align}
    F(u, v) &= \frac{1}{N^2} \sum_{x=0}^{N-1} \sum_{y=0}^{N-1} f(x, y) e^{-\frac{2\pi i}{N}(ux + vy)}, \quad u, v = 0, 1, \dots, N-1, \label{eq:DFT} \\
    f(x, y) &= \sum_{u=0}^{N-1} \sum_{v=0}^{N-1} F(u, v) e^{\frac{2\pi i}{N}(ux + vy)}, \quad x, y = 0, 1, \dots, N-1. \label{eq:iDFT}
\end{align}

We introduce coefficients \( A \) and \( B \) to describe the relationship between the known time-domain sequence and its frequency-domain representation:
\begin{align}
    A &= \sum_{x=0}^{N-1} \sum_{y=0}^{N-1} f(x,y) \left( \frac{e^{-\frac{2\pi i}{N}(ux + vy)}}{N} - \frac{e^{-\frac{2\pi i}{N+1}(ux + vy)}}{N+1} \right), \label{eq:A} \\
    B &= \frac{1}{(N+1)^2} \sum_{x=0}^{N-1} \sum_{y=0}^{N-1} f(x,y) e^{-\frac{2\pi i}{N+1}(ux + vy)}. \label{eq:B}
\end{align}

The new time-domain values \( f(N, y) \) and \( f(x, N) \) can be predicted as:
\begin{align}
    f(N, y) &= (N+1) \left( F'(N, y) - B \right) e^{-\frac{2\pi i}{N+1} N^2}, \label{eq:f_N_y} \\
    f(x, N) &= (N+1) \left( F'(x, N) - B \right) e^{-\frac{2\pi i}{N+1} N^2}. \label{eq:f_x_N}
\end{align}

Similarly, the new frequency-domain values \( F'(u, v) \) are given by:
\begin{align}
    F'(u, v) &= A + \left( F(N+1, v) - B \right) e^{\frac{2\pi i}{N+1}(ux + vy)}, \quad u, v = 0, 1, \dots, N-1. \label{eq:F_prime}
\end{align}

Thus, for each \( u, v \), the new frequency component \( F'(u, v) \) can be inferred from the relationship:
\begin{align}
    F'(u, v) &= A + \left( F'(u, v) - B \right) e^{\frac{2\pi i}{N+1}(ux + vy)}. \label{eq:F_infer}
\end{align}

Once \( F'(u, v) \) is determined, the predicted time-domain values \( f(N, y) \) and \( f(x, N) \) can be obtained by applying the inverse 2D DFT in \eqref{eq:iDFT}.

In conclusion, the 2D FNO predicts the next frequency component \( F'(u, v) \) by using the relationship between time-domain and frequency-domain representations. The coefficients \( A \) and \( B \) are used to infer the new frequency-domain values from the known values \( F(u, v) \). Finally, the inverse DFT transforms \( F'(u, v) \) back to the time domain to obtain the predicted value \( (\hat{x}_N, \hat{y}_N) \). \hfill $\square$

\section{Extra Case Study Result}
\begin{figure*}[h]
  \centering
  \includegraphics[width=\linewidth]{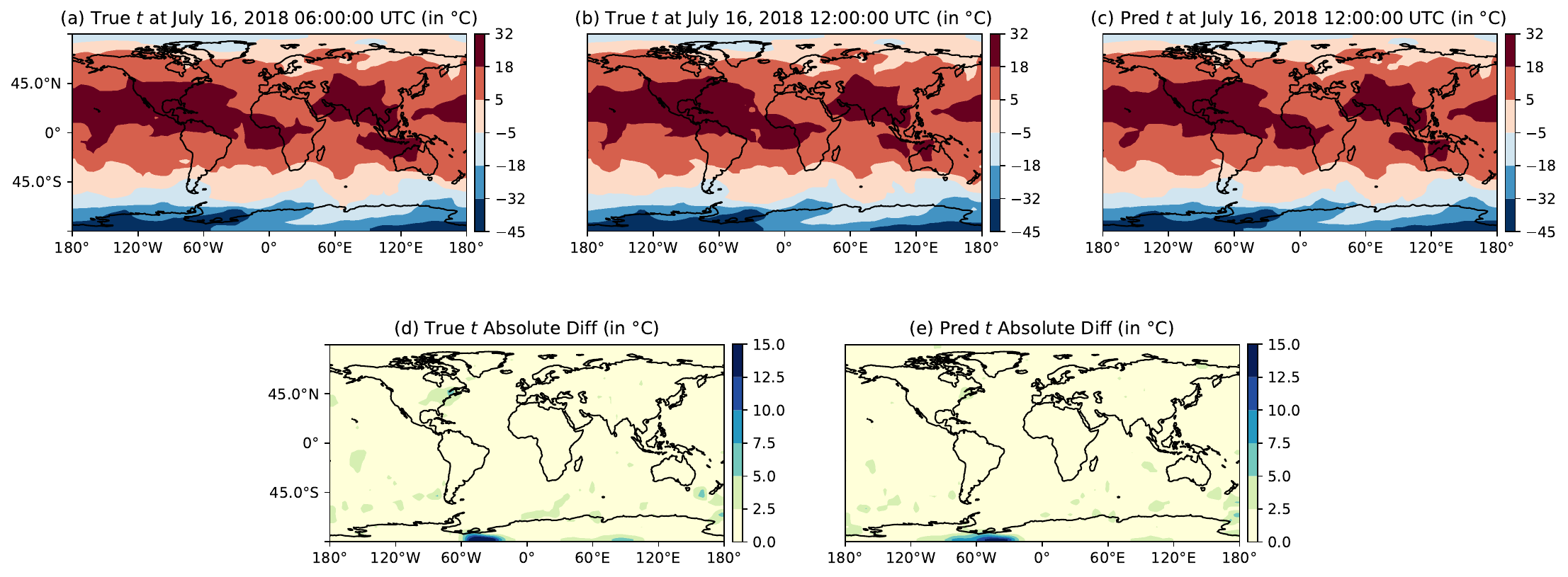}
  \caption{Case Study of variable t}
  \label{ttt}
\end{figure*}
Here in Figure \ref{ttt} we present another case study focusing on the variable t, examining the time period from July 16, 2018 06:00:00 UTC to July 16, 2018 12:00:00 UTC. Our model demonstrates comparable efficacy in capturing these dramatic weather transitions, further validating its robust performance in detecting significant meteorological variations.

\end{document}